\documentclass[conference]{IEEEtran}
\IEEEoverridecommandlockouts
\usepackage{cite}
\usepackage{amsmath,amssymb}
\usepackage{graphicx}
\usepackage{url}

\begin{document}

\title{Extreme Length Generalization in a Compact\\
Recurrent Architecture for One-Shot Exploration}

\author{\IEEEauthorblockN{Izen Thornton\IEEEauthorrefmark{1},
Aaron Shey\IEEEauthorrefmark{1},
William Su\IEEEauthorrefmark{2}}
\IEEEauthorblockA{\IEEEauthorrefmark{1}FRANK Autonomous Systems Inc.,
\url{frankautonomous.com}\\
\IEEEauthorrefmark{2}University of California, Berkeley}}

\maketitle

\begin{abstract}
Autonomous robots on one-shot missions run over horizons far longer than the
trajectories seen during training, under a fixed onboard compute budget. We
present FRANK, a 507K-parameter recurrent
architecture that combines tau-gated recurrent modules, content-addressable
memory, and a feedforward reflex pathway. We evaluate it against recurrent,
state-space, and reduced modular baselines at matched parameter count on four
algorithmic sequence tasks, trained at length 5--20 and evaluated out to two
million tokens. At $100{,}000\times$ the maximum training length, 6 of 10
FRANK seeds retain exactly 100.0\% accuracy, while none of the 50 baseline
configurations does, five architectures at ten seeds each with none left
incomplete (Fisher exact, two-sided $p = 4.2 \times 10^{-6}$). Targeted
lesions across the four tasks yield four distinct component-reliance profiles,
consistent with task-dependent allocation across the recurrent, memory, and
reflex pathways. Separately, a FRANK policy trained in simulation
drives a physical ground vehicle to commanded waypoints through obstacles
without teleoperation.
\end{abstract}

\section{Introduction}

Deep ocean and deep space exploration share a constraint terrestrial robotics
rarely faces: the system has to work the first time, far past the conditions
anyone rehearsed. A controller reliable for the length of a test campaign that
quietly degrades beyond it is the failure a one-shot mission cannot detect in
time.

This is a length generalization problem, and the standard tools fail it in
different ways. Transformers with learned absolute positional embeddings cannot
represent positions past their training range, and relative schemes built for
extrapolation do not rescue the task we study: an ALiBi \cite{alibi} probe
collapses to chance by 100$\times$. Attention's $O(L^2)$ cost is a harder
limit: at two million tokens the score matrix alone would hold roughly
$4 \times 10^{12}$ entries. Recurrent models run at any length but drift:
gated recurrent unit (GRU) \cite{gru} baselines match our model at 10$\times$
and land at 10--30\% at 100{,}000$\times$; Mamba \cite{mamba}, a selective
state-space model, is at chance by 100$\times$ on every seed.

\begin{figure}[t]
\centering
\includegraphics[width=0.70\columnwidth]{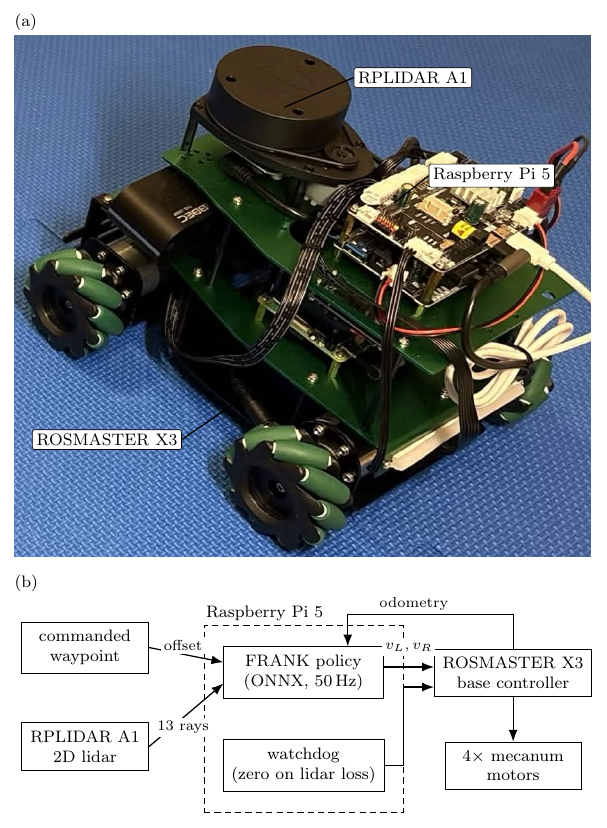}
\caption{System overview. (a) The deployed vehicle. (b) The FRANK policy runs
onboard a Raspberry Pi 5 on a ROSMASTER X3, consuming 13 RPLIDAR A1 rays and
the body-frame waypoint offset and emitting wheel velocities at 50\,Hz; a
watchdog zeroes the command if lidar scans stop.}
\label{fig:hw}
\end{figure}

Prior work reports early results for FRANK \cite{frankpaper1} and a
single-seed lesion analysis \cite{frankpaper2}; the ten-seed re-run with
swept learning rates, the statistical testing, and the hardware deployment
(Fig.~\ref{fig:hw}) are new.

\section{Methodology}

\begin{figure}[!t]
\centering
\includegraphics[width=0.78\columnwidth]{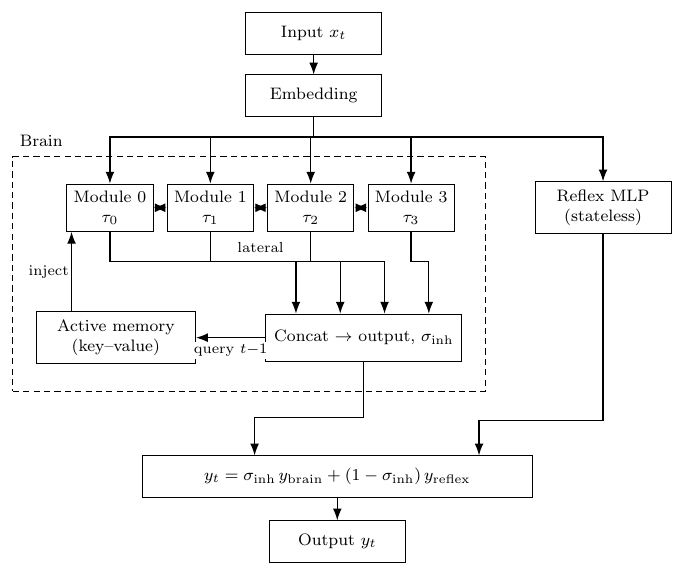}
\caption{The FRANK architecture, after \cite{frankpaper1}. All components
active every timestep; no routing. The inhibit signal blends the recurrent and
reflex output paths.}
\label{fig:arch}
\end{figure}

FRANK (Fig.~\ref{fig:arch}) uses four recurrent modules with learnable time
constants $\tau$ spanning fast to slow, with lateral connections applied after
the module updates. They are simpler than GRUs; there are no multiplicative
gates, only a leaky update
$h_t = (1 - 1/\tau)h_{t-1} + (1/\tau)\tanh(W_x x_t + W_h h_{t-1})$, so a
module's single learned $\tau$, clamped to $[1,100]$ steps, sets how long it
holds information. We call this tau-gated recurrence. A content-addressable
memory holds learned key-value slots queried by the previous hidden state, and
a feedforward reflex pathway provides stateless responses. A veto mechanism
blends the two output paths through a sigmoid inhibit signal from the
recurrent modules, $y = \sigma_{\mathrm{inh}} y_{\mathrm{brain}} +
(1-\sigma_{\mathrm{inh}}) y_{\mathrm{reflex}}$, initialized reflex-dominated.
The configuration holds 507K parameters and baselines are matched to it; a
fifth component in \cite{frankpaper1}, an anomaly detector, was dormant on
every task in \cite{frankpaper2} and is dropped.

All components are active every timestep; nothing selects among them. Because
they differ in computational structure, gradient descent has something to
allocate against. Prior modular architectures route inputs through gating
\cite{moe}, attention \cite{rims}, or a learned router \cite{routing};
Rosenbaum et al. \cite{routing2} found that without a router, homogeneous
modules do not specialize. In their setup that held; the components here are
heterogeneous.

\section{Experimental Evaluation}

\begin{figure}[!tb]
\centering
\includegraphics[width=0.66\columnwidth]{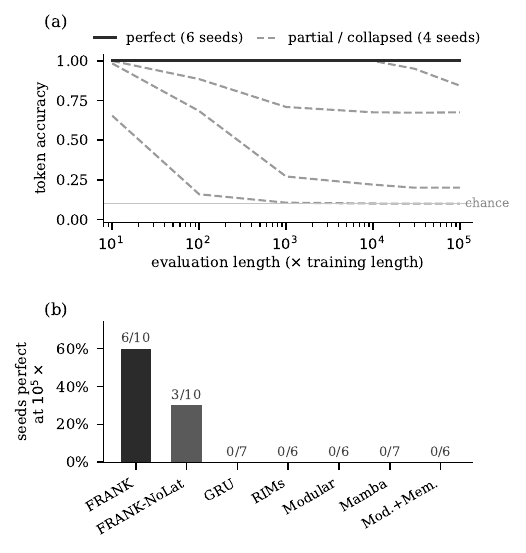}
\caption{(a) Per-seed FRANK accuracy against evaluation scale on the
state-transition task; six seeds are exactly flat at 100\% out to two million
tokens. (b) Seeds reaching perfect accuracy at $10^5\times$ per architecture;
the best baseline seed at that scale reaches 30.3\% (GRU).}
\label{fig:phase2}
\end{figure}

The state-transition task is a running sum modulo 10: inputs are uniformly
random digits $0$--$9$, and the target at each position is the cumulative sum
of every input so far, modulo 10. Ten states, one transition rule, applied at
every step. Data is generated procedurally from fixed seeds, so evaluation
sequences come from the training input distribution and differ only in length.
Each extreme-length cell is ten independent streams, two million tokens each.

We train at lengths 5--20 and evaluate out to 100{,}000$\times$ training
length (Fig.~\ref{fig:phase2}). Six of ten FRANK seeds hold exactly 100.0\% at
every scale. The rest are partial (84.4\%; one stepping at roughly 2{,}700
tokens onto a stable 67.6\% plateau) or failed (20.1\% and chance), and the
failures are structured rather than noisy: each applies a consistent incorrect
rule.

The baselines are GRU, RIMs, and Mamba matched to the 507K configuration, plus
two reduced variants of our own architecture (the recurrent modules alone, and
modules plus memory); the transformer cannot be evaluated at this scale.
Baseline coverage is now complete: all 50 cells, five architectures at ten
seeds each, against FRANK's 10 of 10, and none generalizes
(Fig.~\ref{fig:phase2}b); the best reaches 30.3\%. Fisher's exact test on
seeds gives two-sided $p = 4.2 \times 10^{-6}$. Per stream, 67 of FRANK's 100
two-million-token sequences ran exactly correct end to end against 0 of the
baselines' 500.

Standard-scale testing cannot certify this regime, and the failure is
asymmetric. At 10$\times$ the GRU matches FRANK (96.8\% against 96.6\%).
Weakness at 10$\times$ predicts collapse; strength does not predict success:
of 13 cells at or above 90\% at 10$\times$, seven reach 1.0000 at
100{,}000$\times$ and two fall to chance.

Prior work \cite{frankpaper1} credited the lateral connections with error
correction. The completed ablation does not support that: 6 of 10 against 3 of
10 without laterals is not significant (Fisher $p = 0.37$), and the two
variants' successful seeds share one member, so laterals change which basins
training reaches, not how often it reaches a generalizing one.

\section{Component Lesion Studies}
\label{sec:lesion}

\begin{figure}[!tb]
\centering
\includegraphics[width=0.46\columnwidth]{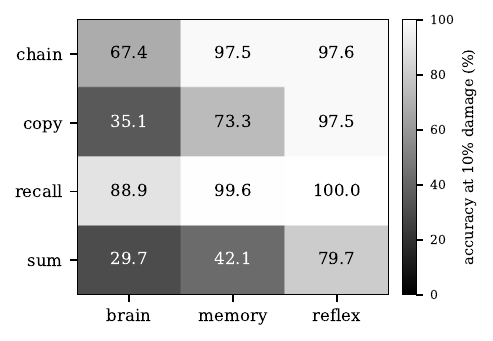}
\caption{Targeted lesion at 10\% damage. Four tasks, four distinct reliance
profiles from the same architecture, with no routing, no task labels, and no
component-specific loss.}
\label{fig:heatmap}
\end{figure}

\begin{figure}[!t]
\centering
\includegraphics[width=0.96\columnwidth]{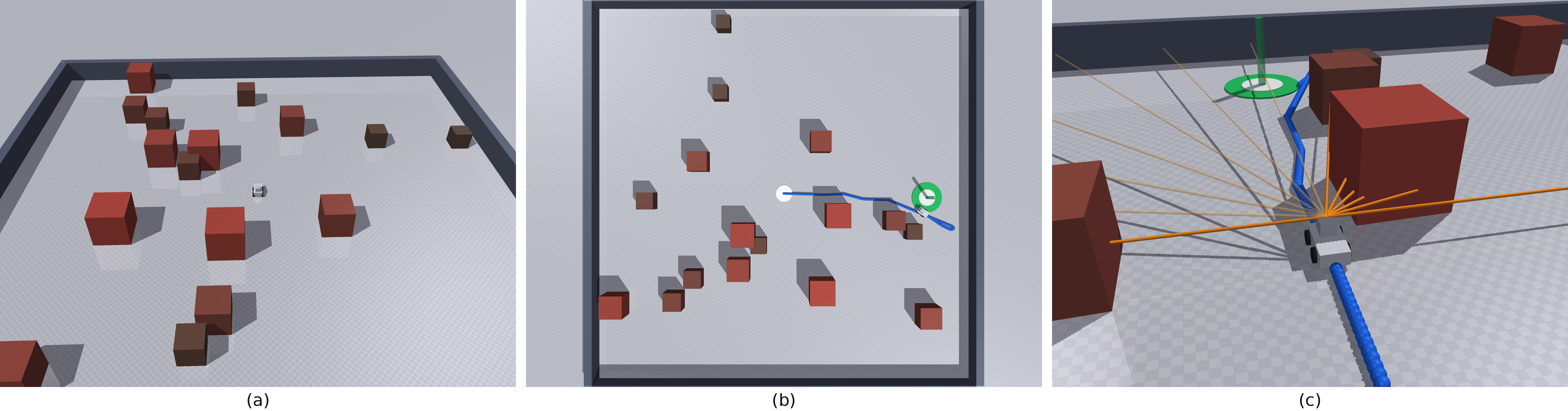}
\caption{Navigation in simulation. (a) The obstacle field used for training.
(b) Commanded waypoint (green) and the executed path. (c) The 13 lidar rays
that form the policy's entire exteroceptive observation.}
\label{fig:sim}
\end{figure}

We train the architecture on four algorithmic sequence tasks and damage one
component at a time, holding the others intact: chain is the state-transition
task above, copy reproduces a token sequence, recall retrieves a token through
intervening noise, and sum maintains a running total. Damage means zeroing a
random fraction of one component's trained weights and re-measuring accuracy,
after the lesion studies in neuroscience the protocol is named for. Every
lesion is scored at the training length of 5--20 tokens, not at extreme
length, so the damage and not the horizon is the variable.
Fig.~\ref{fig:heatmap} gives accuracy at 10\% damage, averaged across seeds
($n=10$ per task; $n=8$ for recall); every cell reaches 97.5\% or better
undamaged, so the damage causes the differences.

No two tasks share a profile. Copy is brain-critical with a reflex path flat
out to 70\% damage. Recall leans on the recurrent modules alone. Chain grades
the components. Sum alone eventually needs all three: its reflex falls to
10.3\% by 70\% damage, where the same component on copy still holds 97.1\%.
Nothing told the model to do this. The inductive biases did the work.

The reflex's low lesion sensitivity and the collapse of the reduced
modules-plus-memory variant in Fig.~\ref{fig:phase2}b sit oddly together,
since that variant is the one without a reflex. Two things differ. The lesion
damages a reflex inside a trained FRANK at training length, where the
recurrent path already carries the task; the reduced variant never had a
reflex, or the veto gate, so training reached a different solution rather than
the same one minus a part. That variant also reaches only 23.8\%
at 10$\times$ against FRANK's 96.6\%, so its failure is not specific to long
horizons. What the reflex contributes to length generalization is untested.

Copy and recall reproduce the single-seed analysis \cite{frankpaper2} almost
exactly, but three of its claims do not: memory is supporting rather than
essential on chain, the reflex is load-bearing late rather than critical early
on sum, and the simultaneous collapse on sum belongs to the full damage
curve.

Under undirected damage the GRU wins outright: at 10\% global weight zeroing
it holds 99.6\% against FRANK's 65.6\%, so a system selected purely for
graceful degradation should use the GRU. That withdraws a further claim, since
\cite{frankpaper2} argued a robustness-generalization inverse resting on FRANK
being the most fragile architecture tested, and FRANK is second most robust
here. What survives is a different signature: across ten random damage
patterns at 20\% chain damage, the standard deviation of FRANK's outcome
(0.0706) is 2--8$\times$ higher than every baseline's (nearest 0.0346), the
highest of the eight architectures in the global-lesion set. That set is one
larger than Fig.~\ref{fig:phase2}b, which shows the seven models evaluable at
extreme length; the transformer is lesioned at training length, where it runs.

\section{Physical Robot Demonstration}

The policy observes 13 lidar rays, the body-frame offset to the commanded
waypoint, and its own previous action over four timesteps, and emits two
numbers at 50\,Hz: normalized left and right wheel velocities. Thirteen rays
are the entire exteroceptive observation (Fig.~\ref{fig:sim}); there is no
camera and no map carried between timesteps, and the offset comes from the
base controller's wheel odometry with no external localization. Steering,
obstacle avoidance, and arrival are all raw network output; there is no planner
or stop gate, and the only behavior outside the policy is a watchdog that
zeroes the command if lidar scans stop. We trained
it in a MuJoCo simulation written for this work on the mjlab framework under
PPO with a staged curriculum and deployed it through ONNX, at larger width
(1.01M parameters). In simulation and on hardware the vehicle reaches the
waypoint while avoiding obstacles without teleoperation, qualitatively at this
stage.

Classical planners remain the right default wherever a map and a reliable pose
estimate exist, and most flight-proven planetary and subsea autonomy is built
that way. The case for a learned policy here is cost, not capability: one
half-million-parameter network, fixed compute per step whatever the horizon,
no replanning and no map, at 50\,Hz on a Raspberry Pi 5. A planetary rover or an AUV on a one-shot mission faces the
same constraint as the sequence tasks above, a fixed onboard budget over a
horizon far past anything rehearsed with no operator to catch slow
degradation, which is why we take the length-generalization result to bear on
those settings. The vehicle stands in for that constraint, not for either
domain's terrain.

\section{Limitations and Future Work}

Generalization is seed-dependent, and the four sequence tasks are synthetic.

Odometry drift is the deployment's binding long-horizon limit, and a different
limit from the one the sequence tasks measure. The waypoint offset is derived
from wheel odometry, which accumulates error without bound on a mecanum base,
so the goal vector the policy consumes degrades even where the policy does
not. The network holds no map and cannot detect the drift, and a one-shot
mission has no ground truth to correct against. Closing that gap is a
state-estimation problem, not a policy one: the same input accepts a pose from
scan matching or any external fix.

The specialization claim lacks a control. The contrast with \cite{routing2} is
drawn against their setup rather than tested here. FRANK with three
structurally identical components, lesioned under this protocol, would settle
it: if the homogeneous variant does not collapse the four profiles, the claim
is wrong. The mechanism is also open; a circular representation of the kind
transformers learn for modular arithmetic was tested on the hidden states and
fails.

Future work runs that control and quantifies the deployment: success rates
across obstacle layouts, path efficiency, failure modes, the sim-to-hardware
gap, degradation against a drifting pose estimate, and onboard latency and
power.

\section{Conclusion}

Six of ten FRANK seeds hold exact accuracy at $10^5\times$ their training
length, where none of the 50 baseline configurations does, and targeted
lesions show the capability sitting in task-dependent places inside one
undifferentiated network.

\end{document}